\documentclass[journal,twoside,web]{ieeecolor}

\makeatletter
\let\NAT@parse\undefined
\makeatother

\usepackage[numbers,sort&compress]{natbib}

\usepackage{generic}
\usepackage{amsmath,amssymb,amsfonts}
\usepackage{graphicx}
\usepackage{algorithm}
\usepackage[noend]{algpseudocode}
\usepackage{textcomp}

\usepackage{array}
\usepackage{pifont}
\usepackage{tabularx}
\usepackage{subcaption}
\usepackage{makecell}
\usepackage[table,dvipsnames]{xcolor}
\usepackage{comment}
\usepackage{afterpage}
\usepackage{booktabs,makecell,siunitx,adjustbox}

\usepackage{hyperref}
\hypersetup{
    colorlinks,
    linkcolor={blue!80!black},
    citecolor={blue!80!black},
    urlcolor={blue!80!black},
    breaklinks=true,
    plainpages=true
}

\newcommand{\cref}[3]{\hyperref[#2]{#1~\ref*{#2}{#3}}}
\newcommand{\crefs}[3]{\hyperref[#2]{#1~\ref*{#2}-\ref*{#3}}}
\newcommand{\colref}[2]{\hyperref[#2]{#1~\ref*{#2}}}

\newcommand{\figref}[1]{\colref{Figure}{#1}}

\newcommand{\tabref}[1]{\colref{Table}{#1}}

\newcommand{\best}[1]{\textbf{#1}}

\def\BibTeX{{\rm B\kern-.05em{\sc i\kern-.025em b}\kern-.08em
    T\kern-.1667em\lower.7ex\hbox{E}\kern-.125emX}}
\begin{document}
\title{In-Context Adaptation of VLMs for Few-Shot Cell Detection in Optical Microscopy}
\author{Shreyan Ganguly, Angona Biswas, Jaydeep Rade, Md Hasibul Hasan Hasib, Nabila Masud, \\Nitish Singla, 
Abhipsa Dash, Ushashi Bhattacharjee, Aditya Balu, \\Anwesha Sarkar, 
Adarsh Krishnamurthy$^*$ and 
Soumik Sarkar$^*$
\thanks{This work was supported in part by the National Science Foundation under grant CPS:2409359}
\thanks{The authors are with Iowa State University, Ames, Iowa, United States. (Correspondence: 
adarsh@iastate.edu; soumiks@iastate.edu)}
}

\maketitle

\begin{abstract}
Foundation vision–language models (VLMs) excel on natural images, but their utility for biomedical microscopy remains underexplored. In this paper, we investigate how in-context learning enables state-of-the-art VLMs to perform few-shot object detection when large annotated datasets are unavailable, as is often the case with microscopic images. We introduce the Micro-OD benchmark, a curated collection of 252 images specifically curated for in-context learning, with bounding-box annotations spanning 11 cell types across four sources, including two in-lab expert-annotated sets. We systematically evaluate eight VLMs under few-shot conditions and compare variants with and without implicit test-time reasoning tokens. We further implement a hybrid Few-Shot Object Detection (FSOD) pipeline that combines a detection head with a VLM-based few-shot classifier, which enhances the few-shot performance of recent VLMs on our benchmark. Across datasets, we observe that zero-shot performance is weak due to the domain gap; however, few-shot support consistently improves detection, with marginal gains achieved after six shots. We observe that models with reasoning tokens are more effective for end-to-end localization, whereas simpler variants are more suitable for classifying pre-localized crops. Our results highlight in-context adaptation as a practical path for microscopy, and our benchmark provides a reproducible testbed for advancing open-vocabulary detection in biomedical imaging.
\end{abstract}


\section{Introduction}
\label{sec:introduction}

\begin{figure*}[t!]
    \centering
    \includegraphics[width=0.78\linewidth, trim={2.25in 0.5in 2.25in 0.5in},clip]{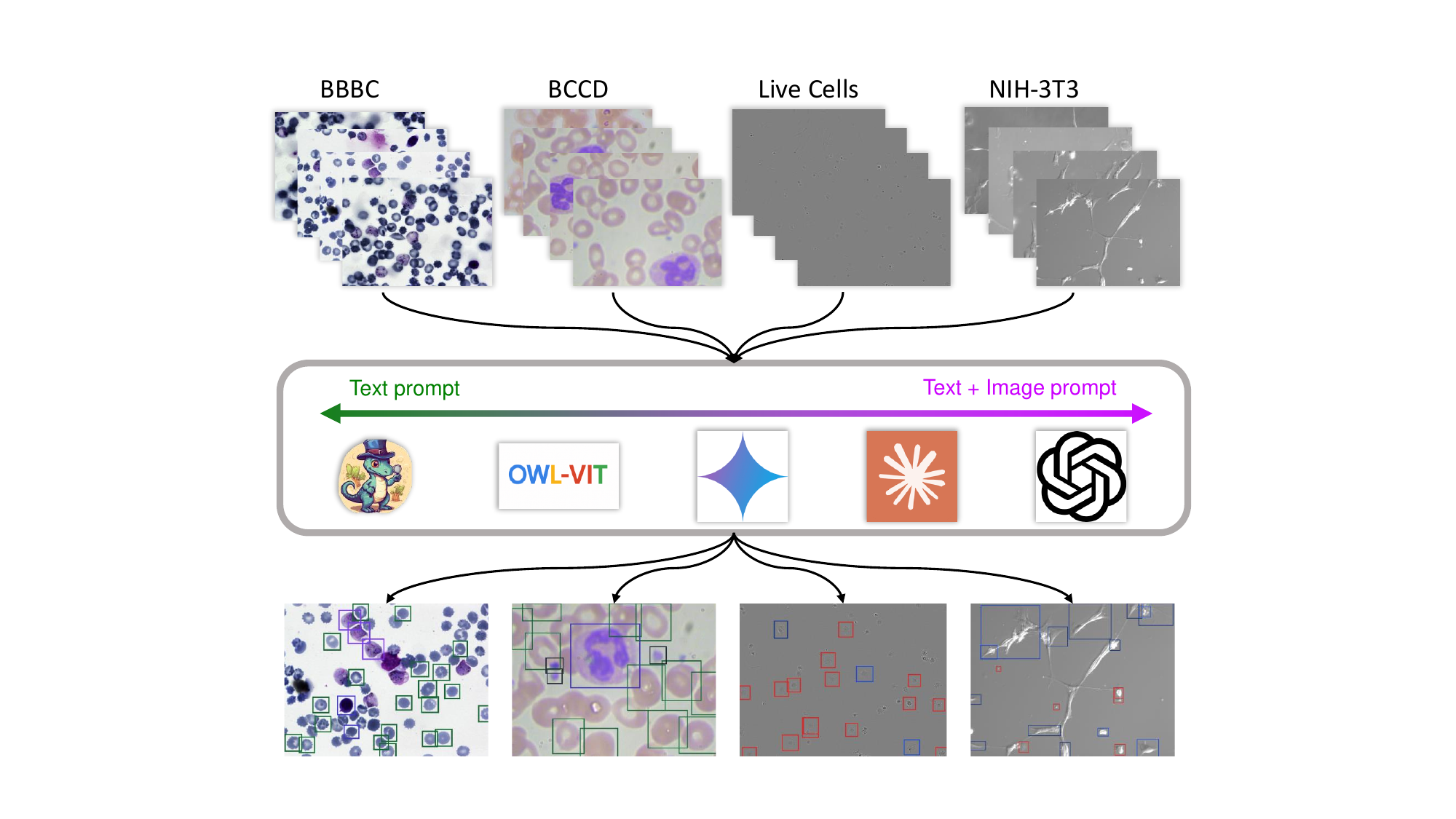}
    \caption{An overview of the Few-Shot Object Detection (FSOD) pipeline using vision-language models (VLM) for cell shape detection. We process four different microscopy datasets (BBBC, BCCD, Live Cells, and NIH-3T3) by various VLMs using either a text-only or a combined text and image prompt to generate bounding box detections for different cell types.}
    \label{fig:overview}
\end{figure*}

\IEEEPARstart{D}{eep} Learning (DL) has revolutionized the field of biomedical imaging by enhancing or automating several imaging-based workflows. This enhancement has spanned diverse modalities, including radiology (CT, MRI, ultrasound) and microscopy (optical microscopy, photoacoustic imaging, and label-free, live-cell imaging)~\citep{zhou2021review}. Among these, microscopy is a cornerstone of modern biology and medicine, providing critical insights into cellular structures and mechanisms essential for disease diagnosis, treatment monitoring, and basic biological discovery~\citep{weissleder2015BioImaging}. However, manual interpretation of microscopy images is time-consuming, subjective, and impractical for large-scale studies such as high-throughput screening or clinical trials~\citep{Gurcan2009pathologyreview}. DL addresses these limitations by automatically extracting hierarchical and fine-grained features from complex, high-dimensional data~\citep{zhou2021review}. This capability enables scalable interpretation of microscopy images for clinical and biological applications. A key example is the detection and localization of individual cells, which supports quantification of morphology, density, and dynamics. These measurements are central to fields such as developmental biology, cancer, immunology, and neuroscience~\citep{caicedo2017data}. High-content microscopy screens further rely on robust cell detection to evaluate treatment effects, toxicity, and drug efficacy at the single-cell level~\citep{smith2022high}. These demands for precision, scalability, and objectivity have driven the adoption of DL-based AI pipelines in microscopy image analysis.

While DL has transformed microscopy image analysis, earlier attempts relied heavily on classical computer vision techniques, such as thresholding, morphology, and template matching, for analyzing cells, lesions, or anatomical structures~\citep {suzuki2017overview}. While effective in restricted settings, these approaches rely on hand-tuned heuristics and degrade when there are shifts in staining, illumination, or instrumentation. Modern, fully supervised deep detectors, such as Faster R-CNN~\citep{ren2016fasterrcnnrealtimeobject} or YOLO~\citep{redmon2018yolov3}, deliver high accuracy when large, high-quality labels are available. Still, they remain constrained by their training dataset, require costly fine-tuning to cover new categories, and are brittle under domain shifts common in microscopy~\citep{zou2023object, litjens2017survey, shen2017deep}. Task-specific training pipelines are expensive (in terms of annotation and computation), slow to adapt (to new cell types/protocols), and semantically closed-world (they cannot recognize unseen classes without retraining).


Few-Shot Object Detection (FSOD)~\citep{Han_2024_CVPR} has offered a pathway to adapt powerful pre-trained models to novel domains with less supervision. While FSOD methods mitigate supervision scarcity, they still depend on complex meta-learning or fine-tuning pipelines and remain limited across various domains. Open-Vocabulary Object Detection (OVD)\cite{minderer2022owlvit} extends this idea by aligning visual features with language, enabling detectors to recognize unseen categories from text prompts without any specific training~\citep{zhang2024vision}. Building on OVD, Foundation models like Grounding DINO~\citep{liu2024grounding} and OWL-ViT~\citep{minderer2022owlvit} exemplify this trend, leveraging large-scale pretraining to localize and classify a wide range of objects described by text prompts. 

The advent of foundation models has revolutionized the paradigm of computer vision, particularly Vision-Language Models (VLMs) that are pre-trained on web-scale multimodal data. By utilizing the extensive prior knowledge embedded in foundation models, it is possible to quickly adapt to new object categories with only a limited number of annotated examples. This process is referred to as in-context learning (ICL)~\citep{zhou2022teaching}.
Generalist models such as GPT~\citep{achiam2023gpt}, Gemini~\citep{comanici2025gemini},
and Claude~\citep{kurokawa2024diagnostic} are trained on enormous datasets of image-text pairs to learn a joint embedding space, where semantically related visual and textual features are mapped close to one another. This alignment enables VLMs to be prompted with natural language descriptions, allowing them to recognize and localize novel categories, even if those categories were not present in the original training data. The use case of these generalized models can be extended to detecting arbitrary objects in photos, leveraging natural language prompting. However, it is observed that the detection performance of these generalist models is confined to images that tend to be more frequent on the internet, and does not generalize to the tail of the distribution of the pretraining data~\citep{robicheaux2025roboflow100vlmultidomainobjectdetection}.

Biomedical microscopy is a relatively underrepresented domain in internet-scale datasets and presents distinctive vision challenges, such as fine-grained visual differences between classes, low contrast, and frequent imaging artifacts~\citep{lozano2024mubenchvisionlanguagebenchmarkmicroscopy}. These factors create a substantial domain shift for models pre-trained on web-scale natural images. The diversity of imaging techniques further widens this gap; for instance, the stained appearance of bright-field microscopy is significantly different from the grayscale, halo-laden images produced by phase-contrast microscopy. These challenges make VLMs especially attractive for biomedical microscopy, where new cell types or imaging conditions often fall outside any pretraining distribution. Yet, despite their promise of prompt-driven generalization, their effectiveness in this domain remains underexplored. To address this gap, we introduce Micro-OD, a curated benchmark for microscopy-specific evaluation under in-context learning, and conduct a systematic study. An overview of the FSOD pipeline and evaluation workflow is shown in \figref{fig:overview}.
Our main contributions include:
\begin{itemize}
    \item Micro-OD, a curated benchmark for few-shot object detection in microscopy. It comprises 252 images with bounding-box annotations for 11 cell types, sourced from public and in-house datasets (BCCD~\citep{BCCD_Dataset}, BBBC~\citep{BBBC041v1}, LIVECell C6~\citep{LIVECell_RatC6}, and NIH-3T3).
    \item Comprehensive evaluation of eight state-of-the-art VLMs under zero-shot and few-shot (K=1,3,6) conditions. Our analysis uniquely compares variants with and without implicit reasoning ("thinking") tokens to assess their impact on localization and classification tasks.
    \item Validation of a hybrid FSOD pipeline that decouples localization from classification. This cascaded approach proves highly effective, significantly outperforming end-to-end methods and reaching a top mF1 score of 0.30, a fivefold increase over the best zero-shot baseline.
\end{itemize}
Together, these contributions provide the first comprehensive study of VLMs for few-shot cell detection in microscopy, offering insights into the limitations and opportunities of adapting foundation models to biomedical domains.

\section{Related Work}
\label{sec:related_works}

\subsection*{Vision Language Foundation Models for Object Detection}
Fully supervised object detectors perform well in closed-set scenarios but struggle to recognize categories beyond those seen during training. Open-vocabulary object detection (OVD) addresses this limitation by coupling visual features with natural language cues. Through large-scale multi-modal pretraining, images and text are aligned in a shared embedding space, enabling detectors to localize and classify objects described by text prompts.

Early OVD systems distilled knowledge from frozen \textsc{CLIP} encoders into two-stage detectors~\citep{hinton2015distillingknowledgeneuralnetwork}. For example, ViLD~\citep{gu2022openvocabularyobjectdetectionvision} demonstrated that Mask R-CNN~\citep{he2018maskrcnn} trained with vision–language distillation can localize novel LVIS categories~\citep{gupta2019lvisdatasetlargevocabulary}. GLIP~\citep{li2022groundedlanguageimagepretraining} advanced this idea by jointly training phrase grounding and detection on 27M image-text pairs, producing representations that natively capture region-language alignment. Transformer-based models subsequently built on this foundation: OWL-ViT~\citep{minderer2022owlvit} pre-trains a vanilla ViT~\citep{dosovitskiy2021imageworth16x16words} and fine-tunes it end-to-end for open-world localization, reaching performance on par with task-specific detectors. OV-DETR~\citep{Zang_2022} incorporates conditional matching to handle arbitrary queries, and Grounding DINO~\citep{liu2024grounding} introduces a cross-modality decoder that integrates text and visual cues, achieving substantial gains in open-vocabulary localization. MQ-Det~\citep{xu2023multimodalqueriedobjectdetection} extends GLIP with a gated adapter that accepts both text and image queries, and OWL-ViT~\citep{minderer2022owlvit} can also be adapted to support image-guided detection.

\subsection*{Few-Shot Adaptation of Vision Language Models}
Although foundation VLMs exhibit strong generalization, adapting them to domain-specific tasks using a supervised fine-tuning approach is often challenging due to the scarcity of labeled data. \cite{madan2024revisitingfewshotobjectdetection} showed that, under a stricter K-shot evaluation protocol aligned to application semantics, powerful open-vocabulary localizers such as Grounding DINO can already surpass several classical few-shot detectors without fine-tuning. This highlights the strength of pretrained cross-modal alignment for localization under sparse supervision. Furthermore, \cite{Han_2024_CVPR} proposed FM-FSOD, which leverages modern vision foundation models such as DINOv2~\citep{oquab2024dinov2} as the visual backbone and employs an LLM to reason over query image proposals using contextualized few-shot support images. This approach outperformed conventional learned heads on VOC/COCO while requiring minimal task-specific tuning, suggesting that textual reasoning can substitute for heavy detector training when labeled examples are scarce.


Instance-level personalization has also been explored in IPLoc~\citep{doveh2025teachingvlmslocalizespecific}, which fine-tunes a VLM with dialogue-style supervision on a small set of annotated images, each containing a category label and bounding box. The model is then tasked with localizing the same object type in a query image. This approach enables reliable localization of user-specified objects while preserving general capabilities, providing a practical pathway for on-the-fly, user-guided adaptation when target appearances diverge from generic class descriptions. These approaches demonstrate the versatility of VLMs with few-shot samples, but they have primarily been evaluated on natural images and general-purpose domains. Their effectiveness in specialized settings such as microscopy, however, remains unclear. To address this gap, our study focuses on the microscopy domain. It analyzes when in-context examples are most effective, how many in-context few-shot support samples are needed, and how they should be allocated between localization and classification under severe domain shift.

\subsection*{Few-Shot Detection in Biomedical \& Microscopic Images}
Microscopic imagery poses unique challenges, such as objects of interest that are often small, densely packed, low in contrast, and subject to modality-specific artifacts, which frequently degrade the performance of detectors trained on natural images. \citet{rade2022cellshape} trained YOLOv3~\citep{redmon2018yolov3} using 114 annotated phase-contrast images to localize round, spindle, and polygonal live-cell morphologies, enabling automated region selection for downstream analysis. Early transformer-based efforts such as Cell-DETR~\citep{Prangemeier_2020} adapted DETR~\citep{carion2020DETER} for yeast-cell instance segmentation and detection, demonstrating the feasibility of set prediction in biomedical settings. More recently, FSDAOD~\citep{inayat2024fewshotdomainadaptiveobject} addressed cross-domain adaptation with only a few target shots, achieving state-of-the-art performance on two public cell datasets through class sampling and inter/intra-domain feature alignment. \citet{verma2024vlmMicroscopicImages} evaluated the performance of several VLMs, including GPT, LLaVA~\citep{liu2023visualinstructiontuning}, and Gemini, as well as the Segment Anything Model (SAM)~\citep{kirillov2023segment}, on various tasks (classification, segmentation, counting, and visual-question answering) involving microscopy images. They found that while models like GPT and Gemini show promise in understanding visual features in microscopy images, their performance is not yet comparable to that of a human domain expert. They are easily challenged by common complexities in these images, such as impurities, defects, and overlapping artifacts. To provide a standardized benchmark, \mbox{$\mu$-Bench}~\citep{lozano2024mubenchvisionlanguagebenchmarkmicroscopy} compiled 22 biomedical perception tasks, including object detection, across electron, fluorescence, and light microscopy, and showed that current VLMs struggle even on biologically simple cases such as images containing only the target object, where basic foreground–background segmentation would generally suffice. Complementing this, ~\citep{robicheaux2025roboflow100vlmultidomainobjectdetection} introduced Roboflow100-VL, a large-scale benchmark highlighting the poor generalization of VLMs to out-of-distribution domains, particularly in medical imaging under zero-shot and few-shot conditions.

Although VLMs can be adapted with only a few examples, their robustness in microscopic imaging remains limited. In this work, we address this gap by applying few-shot VLM adaptation to diverse microscopy images spanning different biological cell types and by systematically evaluating the object detection task.

\section{FSOD Benchmark}
\label{sec:benchmark}
We introduce Micro-OD, a benchmark designed to evaluate Vision–Language Models for few-shot cell detection and classification in microscopy. Unlike prior benchmarks that sample broadly from dissimilar classes~\citep{bennequin2022fewshotimageclassificationbenchmarks}, Micro-OD targets fine-grained visual localization where cell types share high semantic and morphological similarity. The benchmark integrates three public datasets along with an in-lab curated dataset, spanning diverse imaging modalities and experimental conditions. By providing a curated, multi-source dataset with dedicated few-shot splits, Micro-OD helps in an evaluation framework that addresses the lack of standardized resources for few-shot evaluation in medical imaging.

\begin{table*}[t!]
\centering
\caption{Test and Example split counts for the Micro-OD benchmark. This table details the distribution of annotated cell counts across the four datasets (BCCD, BBBC, NIH-3T3, and LIVECell) and their respective test and reference splits.}
\label{tab:cell_splits_grid}
\setlength{\tabcolsep}{5pt}
\renewcommand{\arraystretch}{1.1}
\footnotesize

\begin{subtable}[t]{.48\textwidth}
\centering
\caption{\textcolor{blue}{BCCD} (364 images; 53 Test, 10 Example)}
\begin{tabular}{lcc}
\toprule
Class & Ref & Sup\\
\midrule
Platelets & 159 & 10\\
Red Blood Cells & 737 & 58\\
White Blood Cells & 56 & 10\\
\bottomrule
\end{tabular}
\end{subtable}\hfill
\begin{subtable}[t]{.48\textwidth}
\centering
\caption{\textcolor{blue}{BBBC} (1328 images; 53 Test, 10 Example)}
\begin{tabular}{lcc}
\toprule
Class & Ref & Sup\\
\midrule
Gametocyte Cells & 24 & 6\\
Red Blood Cells  & 3690 & 684\\
Ring Cells       & 34 & 6\\
Schizont Cells   & 10 & 6\\
Trophozoite Cells& 193 & 26\\
White Blood Cells& 49 & 6\\
\bottomrule
\end{tabular}
\end{subtable}

\medskip

\begin{subtable}[t]{.48\textwidth}
\centering
\caption{\textcolor{blue}{NIH-3T3} (63 images; 53 Test, 10 Example)}
\begin{tabular}{lcc}
\toprule
Class & Ref & Sup\\
\midrule
Polygonal Cells & 303 & 43\\
Round Cells     & 11 & 6\\
Spindle Cells   & 62 & 13\\
\bottomrule
\end{tabular}
\end{subtable}\hfill
\begin{subtable}[t]{.48\textwidth}
\centering
\caption{\textcolor{blue}{LIVECell} (420 images; 53 Test, 10 Example)}
\begin{tabular}{lcc}
\toprule
Class & Ref & Sup\\
\midrule
Polygonal Cells & 114 & 15\\
Round Cells     & 13 & 6\\
Spindle Cells   & 96 & 19\\
\bottomrule
\end{tabular}
\end{subtable}

\medskip
\begin{tabular}{lcc}
\toprule
\textbf{TOTAL CELLS}  & \textbf{5551} & \textbf{914} \\
\textbf{TOTAL CELL SLIDES} & \textbf{212}  & \textbf{40}  \\
\bottomrule
\end{tabular}
\end{table*}

\subsection*{Datasets}

\textbf{Blood Cell Count Dataset (BCCD)}~\citep{BCCD_Dataset}: A publicly available dataset with 364 bright-field microscopy images annotated as the three blood cell categories of Platelets, Red Blood Cells (RBCs), and White Blood Cells (WBCs). This dataset is intensely used as a benchmark for cell detection and has bounding-box level annotations for ground-truth evaluation.

\textbf{Malaria (Plasmodium Vivax) (Broad Bioimage Benchmark Collection - BBBC) Dataset}~\citep{BBBC041v1} – A publicly available collection that has 1328 bright-field microscopy images of thin blood smears annotated into multiple parasitic stages (e.g., Gametocyte, Ring, Schizont, Trophozoite; with RBCs and WBCs). This dataset captures the morphological variation of cells and serves as a good testbed for characterizing fine-grained cell classification.

\textbf{Rat C6 (LIVECell subset) Dataset }~\citep{LIVECell_RatC6}: A publicly available phase-contrast microscopy dataset with 456 images of the rat C6 glioma cell line. The original LIVECell dataset is unguided in shape-based annotations, so we re-annotated the images into three morphological categories: Polygonal, Round, and Spindle cells. Because C6 cells exhibit irregular and heterogeneous morphologies, this dataset is useful for testing shape-based generalization.

\textbf{NIH 3T3 Dataset (BACM lab collection, Iowa State University)} - Beyond publicly available sources, we also put together a small in-lab collection of 63 phase-contrast microscopy images of the NIH 3T3 fibroblast cell line. These images were collected using the camera system attached to an atomic force microscopy (AFM) platform. The AFM employed two cameras: a top camera (640 × 480 resolution) and a bottom camera (1388 × 1040 resolution), as well as an additional level of optical zoom at 20$\times$.

\subsection*{Micro-OD Benchmark}
Curating the dataset, we sample 63 slides from each source dataset with the largest number of annotated boxes, creating a pool of 252 candidate images. We divide this pool into two disjoint splits - ``test'' and ``example'' - using a two-phase Integer Linear Programme (ILP). For each source dataset, we select 53 test slides and 10 example slides, as summarized in \tabref{tab:cell_splits_grid}.

The optimiser enforces two hard constraints on every cell class: at least six boxes in the example split and at least ten in the test split. Phase~1 maximises class-image coverage in the example set; Phase~2 then minimises the surplus boxes of frequent classes in the example while leaving as many weighted boxes as possible for the test. To evaluate the quality of the example split, we compute the \emph{Support-Spread Score} (SSS), which combines class coverage and balance:
\begin{equation}
\text{SSS} = \text{CPC} \times \text{CBE},
\end{equation}
where Class-Presence Coverage (CPC) measures the fraction of images per class included in the example split, and Class Balance Entropy (CBE) is the normalised entropy of the class distribution. We rerun the ILP with 1000 shuffled seeds and retain the split with the highest SSS. Our final split achieves an SSS of 0.74 (coverage 0.83, balance 0.89), indicating both high diversity and good class balance in the few-shot examples.

\begin{algorithm}[t]
\caption{Few-Shot Dataset Split via Two-Phase ILP}
\label{alg:split}
\begin{algorithmic}[1]

\Statex
\State Two disjoint splits: \textbf{Test} and \textbf{Example}.

\Statex
\State \textbf{Variables:} $x_i^{\text{exp}}, x_i^{\text{test}} \in \{0,1\}$ (select image $i$).

\Statex
\State \textbf{Constraints:}
\State \hspace{\algorithmicindent} For each class $c$: at least $m_{\text{exp}}$ boxes in example and $m_{\text{test}}$ in test.
\State \hspace{\algorithmicindent} $x_i^{\text{exp}} + x_i^{\text{test}} \le 1$ for all $i$.
\State \hspace{\algorithmicindent} Split sizes fixed: $\sum_i x_i^{\text{exp}} = n_{\text{exp}}$, $\sum_i x_i^{\text{test}} = n_{\text{test}}$.

\Statex
\State \textbf{Phase 1 (Coverage):}
\State Maximize class–image coverage in support by choosing images that cover as many classes as possible.

\Statex
\State \textbf{Phase 2 (Balancing):}
\State Refine the split to avoid over-represented classes in support and prioritize rarer classes in reference.

\Statex
\State \textbf{Evaluation (per trial):}
\State On the resulting support split, compute:
\State \hspace{\algorithmicindent} (1) \emph{Class–Presence Coverage (CPC)} = average fraction of support images containing each class.
\State \hspace{\algorithmicindent} (2) \emph{Class–Balance Entropy (CBE)} = entropy of the class distribution, normalized by classes.
\State Define $\text{SSS} = \text{CPC} \times \text{CBE}$.

\Statex
\State \textbf{Seed search:}
\State Repeat Phases~1–2 with $T$ seeds; select the split with the highest SSS.
\end{algorithmic}
\end{algorithm}

\section{Benchmark Evaluation Strategy}
\label{sec:exp_setup}

\begin{figure*}[t!]
    \centering
    \includegraphics[width=0.99\linewidth,, trim={0in 0in 0in 0.2in},clip]{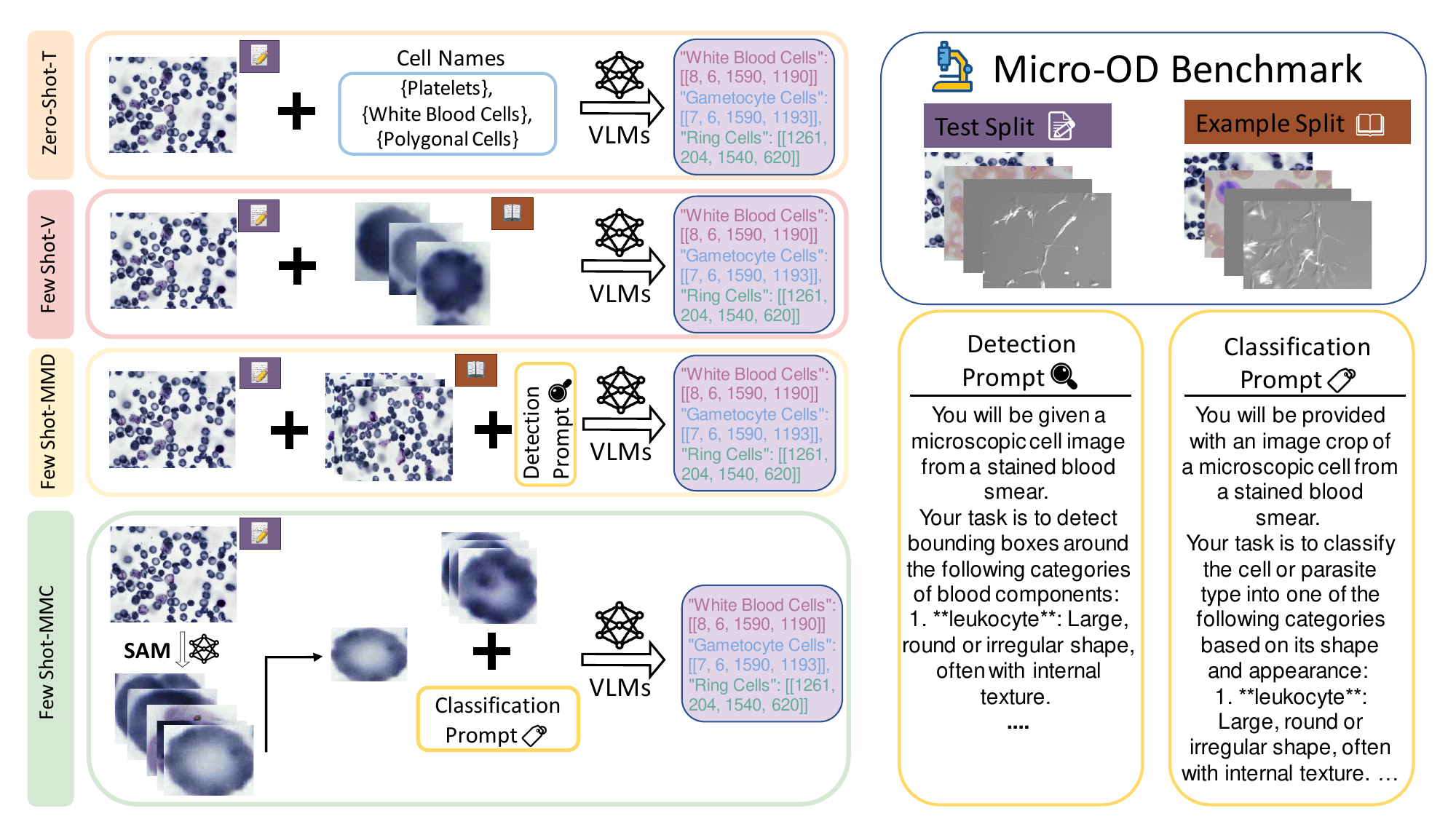}
    \caption{The four experimental configurations used to evaluate Vision-Language Models (VLMs) on the Micro-OD benchmark. This figure illustrates the distinct setups: Zero Shot-T (text-only prompt), Few Shot-V (visual-only prompts), Few Shot-MMO (multi-modal few-shot detection), and Few Shot-MMC (a cascaded pipeline of localization with SAM followed by few-shot classification with VLMs).}
    \label{fig:VLM_FSOD}
\end{figure*}

To systematically evaluate the proposed benchmark, we consider a diverse set of multimodal models, ranging from foundation detectors such as OWL-ViT~\citep{minderer2022owlvit} and Grounding DINO~\citep{liu2024grounding} to proprietary large-scale VLMs including GPT-4~\citep{achiam2023gpt}, Gemini-2.5~\citep{comanici2025gemini}, Claude-3.7~\citep{kurokawa2024diagnostic}, and their respective variants. We designed four experiments with the aim of determining the capabilities of these models in localizing and classifying objects using different input-output configurations. 

We outline the experimental planning methods below and provide descriptions of the four experimental configurations, as well as the role of few-shot adaptation in understanding model flexibility and robustness. \figref{fig:VLM_FSOD} provides an overview of the experimental configurations and their respective settings. The \textbf{output} consists of the bounding box coordinates for all experiments.

\noindent\textbf{1. Experiment Zero Shot-T} 

\noindent \textbf{Input:} Text Prompt, Test Image \\
\textbf{Models:} OWL-ViT, Grounding DINO, Gemini-2.5-Flash, Gemini-2.5-Flash-Thinking, Claude-3.7-Sonnet, Claude-3.7-Sonnet-Thinking

In this baseline experimental condition, models were queried with pure text prompts that describe the target object classes, without visual examples (true zero-shot condition). This condition evaluated the models' open-vocabulary detection capabilities by asking them to predict bounding boxes and predictions based on class names without demonstration or fine-tuning context. The experiment was focused on zero-shot detection. 

\noindent\textbf{2. Experiment Few Shot-V}

\noindent \textbf{Input:} Image Prompt, Test Image\\
\textbf{Models:} OWL-ViT

This experiment used OWL-ViT in a for detection using only visual context in a few-shot setting. In each instance, the model was presented with annotated example images containing both object crops and classes. In few-shot scenarios, 1, 3, or 6 object crops were provided with test image as example image prompts. This setup aimed to evaluate how well localization and classification performances could be improved through visual demonstrations alone. 

\noindent\textbf{3. Experiment Few-Shot-MMD}

\noindent \textbf{Input:} Text Prompt, Image Prompt, Test Image\\
\textbf{Models:} OWL-ViT, GPT-4o, GPT-5, Gemini-2.5-Flash, Gemini-2.5-Flash-Thinking, Claude-3.7-Sonnet, Claude-3.7-Sonnet-Thinking

In this experiment, we used state-of-the-art VLMs and tried to leverage ability for joint vision-language reasoning for object detection. For each detection task in API call, the model was given the full image and text prompt describing the object classes, and 1, 3, or 6 demonstration examples in the prompt (few-shot). This investigated the accuracy and reliability of bounding box predictions and class assignments are impacted by incremental multi-modal examples.

\noindent\textbf{4. Experiment Few-Shot-MMC}

\noindent \textbf{Input:} Text Prompt, Image Prompt, Test Image\\
\textbf{Models:} OWL-ViT, GPT-4o, GPT-5, Gemini-2.5-Flash, Gemini-2.5-Flash-Thinking, Claude-3.7-Sonnet, Claude-3.7-Sonnet-Thinking

In the final experiment, object localization was conducted with the SAM (Segment Anything Model) \cite{kirillov2023segment} to crop out object areas from the full images. The predicted cropped areas and classification prompt as text with 1, 3, or 6 demonstration examples were passed to the state-of-the-art generative VLMS for classification. This cascaded setup evaluated the models’ ability to assign accurate class labels to object areas when provided with explicit textual demonstrations, while specifically evaluating the impact of few-shot text-based support.

\section{Results}
\label{sec:exps}

We conduct the discussed experiments with state-of-the-art Vision-Language models on our microscopic benchmark and report the overall result in this section.

\subsection{Metrics}
Our primary evaluation metric is the mean F1 score ($mF1$), which provides a comprehensive measure of detection accuracy. To compute this score, we first match predicted bounding boxes to ground truth boxes using the Hungarian algorithm, a method inspired by DETR\cite{carion2020endtoendobjectdetectiontransformers} that we chose over the standard greedy approach.

\begin{table}[b!]
\centering
\caption{Evaluation results for the few-shot object detection in a zero-shot setting. We report the mF1 score and Mean IoU (TP@0.5) for various VLMs.}
\label{tab:results_k0}
\footnotesize

\begin{adjustbox}{max width=\textwidth}
\begin{tabular}{l *{2}{S[table-format=1.2]}}
\toprule
\makecell[l]{\textbf{Method}} &
\multicolumn{1}{c}{mF1[.05:.70]$\uparrow$} &
\multicolumn{1}{c}{\makecell{Mean IoU\\(TP@0.5)$\uparrow$}} \\
\midrule
\rowcolor{gray!15}\multicolumn{3}{l}{\textbf{EXP1: Zero Shot-T}} \\
GroundingDINO & 0.04 & 0.53 \\
OWL-ViT       & 0.03 & 0.40 \\
Gemini-2.5-Flash & 0.05 & 0.57 \\
Gemini-2.5-Flash-Thinking & \best{0.06} & \best{0.59} \\
Claude-3.7-Sonnet & 0.03 & 0.34\\
Claude-3.7-Sonnet-Thinking & 0.03 & 0.31\\
\bottomrule
\end{tabular}
\end{adjustbox}
\end{table}

\begin{table*}[t]
\centering
\caption{Evaluation results for the few-shot object detection in a few-shot setting. We compare the performance of different models across three few-shot conditions (K=1,3,6) for the Few Shot-V, Few Shot-MMO, and Few Shot-MMC experiments.}
\label{tab:results_k136}
\footnotesize

\begin{adjustbox}{max width=\textwidth}
\begin{tabular}{
  l
  *{2}{S[table-format=1.2]}
  *{2}{S[table-format=1.2]}
  *{2}{S[table-format=1.2]}
}
\toprule
& \multicolumn{2}{c}{\textbf{K=1}} & \multicolumn{2}{c}{\textbf{K=3}} & \multicolumn{2}{c}{\textbf{K=6}} \\
\cmidrule(lr){2-3}\cmidrule(lr){4-5}\cmidrule(lr){6-7}
\makecell[l]{\textbf{Method}} &
{mF1[.05:.70]$\uparrow$} & {\makecell{Mean IoU\\(TP@0.5)$\uparrow$}} &
{mF1[.05:.70]$\uparrow$} & {\makecell{Mean IoU\\(TP@0.5)$\uparrow$}} &
{mF1[.05:.70]$\uparrow$} & {\makecell{Mean IoU\\(TP@0.5)$\uparrow$}} \\
\midrule
\rowcolor{gray!15}\multicolumn{7}{l}{\textbf{Few Shot-V}} \\
OWL-ViT                  & \best{0.08} & \best{0.77} & \best{0.11} & \best{0.79} & \best{0.18} & \best{0.81} \\
\addlinespace[2pt]
\rowcolor{gray!15}\multicolumn{7}{l}{\textbf{Few Shot-MMD}} \\
GPT-4o                  & 0.05 & 0.58 & 0.05 & 0.59 & 0.05 & 0.56 \\
GPT-5                   & 0.07 & 0.50 & 0.08 & 0.56 & 0.08 & 0.55 \\
Gemini-2.5-Flash        & 0.04 & 0.47 & 0.05 & 0.58 & 0.04 & 0.48 \\
Gemini-2.5-Flash-Thinking & \best{0.17} & \best{0.71} & \best{0.21} & \best{0.72} & \best{0.21} & \best{0.72} \\
Claude-3.7-Sonnet       & 0.03 & 0.13 & 0.03 & 0.12 & 0.03 & 0.43 \\
Claude-3.7-Sonnet-Thinking & 0.04 & 0.39 & 0.04 & 0.29 & 0.04 & 0.43 \\
\addlinespace[2pt]
\rowcolor{gray!15}\multicolumn{7}{l}{\textbf{Few Shot-MMC}} \\
GPT-4o                   & \best{0.24} & \best{0.83} & \best{0.28} & \best{0.83} & \best{0.30} & \best{0.84} \\
GPT-5                    & \best{0.26} & \best{0.79} & \best{0.27} & \best{0.78} & \best{0.28} & \best{0.78} \\
Gemini-2.5-Flash         & 0.20 & 0.79 & 0.22 & 0.79 & 0.22 & 0.79 \\
Gemini-2.5-Flash-Thinking & 0.12 & 0.61 & 0.18 & 0.61 & 0.17 & 0.56 \\
Claude-3.7               & 0.08 & 0.80 & 0.12 & 0.80 & 0.11 & 0.80 \\
Claude-3.7-Thinking      & 0.07 & 0.79 & 0.13 & 0.80 & 0.11 & 0.80 \\
\bottomrule
\end{tabular}
\end{adjustbox}
\end{table*}

We then calculate precision and recall at each of 50 distinct Intersection over Union ($IoU$) thresholds, ranging from $0.05$ to $0.70$. The F1 score for a given threshold is the harmonic mean of precision and recall:
$$ F1 = \frac{2 \times \text{Precision} \times \text{Recall}}{\text{Precision} + \text{Recall}} $$

The final $mF1$ is the mean of the F1 scores obtained across these 50 $IoU$ thresholds:
$$ mF1 = \frac{\sum_{IoU=0.05}^{0.70} F1@IoU}{50}$$

We choose to report this aggregated score because detecting cells in microscopic images is a challenging task due to their rare representation in the pre-training sets of these models. Evaluating across a spectrum of $IoU$ thresholds reflects how robust and accurate the model's predictions are under varying localization criteria.

In addition to $mF1$, we also report the Mean IoU of true positives at $IoU=0.5$ (Mean IoU (TP@0.5)). 
This metric captures the average overlap of all predicted boxes that were successfully matched to a ground truth instance at the $IoU=0.5$ threshold, which is a standard operating point in object detection benchmarks. 
Formally, for each true positive match at $IoU \geq 0.5$, we compute its IoU with the ground truth box; the final score is the arithmetic mean across all such matches. If no true positives are found at this threshold, the metric is defined as $0$. This measure complements $mF1$ by providing a more fine-grained indication of how well the matched detections align spatially with ground-truth objects, rather than just whether they exceeded the threshold. 
While $mF1$ reflects a model’s balance of precision and recall across multiple overlap criteria, Mean IoU (TP@0.5) directly quantifies the localization accuracy of the correctly detected objects.

\subsection{Zero-Shot Detection}
Under true zero-shot conditions, where vision foundation detectors were given only text-based prompts, both models performed poorly. This reflects the significant domain gap between their natural-image pre-training and the specialized characteristics of microscopy. We evaluate the zero-shot performance of six Vision Language Models and report their scores in \tabref{tab:results_k0}. Grounding DINO achieved an $mF1$ score of $0.04$ with a Mean $IoU$ of $0.53$, while OWL-ViT scored an $mF1$ of $0.03$ with a Mean $IoU$ of $0.40$. Among the new models, Gemini-2.5-Flash and its thinking variant performed better, with the latter achieving $mF1$ of $0.06$ and Mean IoU of $0.59$. In contrast, Claude-3.7-Sonnet and its thinking variant showed lower performance, scoring $mF1$ $0.03$. These results indicate that while the detectors can occasionally place bounding boxes with reasonable overlap (as shown by the moderate Mean $IoU$), their overall detection quality across a range of $IoU$ thresholds is low without visual examples.

\subsection{Few-Shot Detection}
We next evaluate- the models in few-shot settings, providing $K=1$, $3$, and $6$ examples for in-context learning. We report the corresponding scores in \tabref{tab:results_k136}.

\textbf{Experiment Few Shot-V}: Providing OWL-ViT with only visual examples (cell crops) was ineffective overall. For $K = 1, 3,$ and $6$, the $mF1$ score improved steadily from $0.08$ to $0.11$ to $0.18$, and the Mean $IoU$ increased from $0.77$ to $0.79$ to $0.81$, respectively. This demonstrates that visual context alone can somewhat help bridge the domain gap, even without accompanying instructional text.

\textbf{Experiment Few Shot-MMO}: Adding multimodal examples significantly improved detection performance, with the "thinking" model variants showing notable gains. As the number of examples ($K$) increased from 1 to 3, Gemini-2.5-Flash-Thinking's $mF1$ score rose from $0.17$ to $0.21$, however it saturated, maintaining Mean $IoU$ around $0.71$. This performance substantially exceeded that of non-thinking models like GPT-5 ($mF1$ of $0.08$) and GPT-4o ($mF1$ of $0.05$). The modest gains from $K=1$ to $K=6$ suggest that performance may saturate after only a few demonstrations are provided.

\textbf{Experiment Few Shot-MMC}: In this setup, object localization was first performed by SAM, and VLMs were then tasked with classifying the resulting crops. Here, the non-thinking GPT-family models led in performance. GPT-4o's $mF1$ score increased from $0.24$ to $0.28$ to $0.30$ as $K$ increased, with a consistently high Mean $IoU$ of approximately $0.83$--$0.84$. Similarly, GPT-5's $mF1$ score went from $0.26$ to $0.27$ to $0.28$. In contrast, Gemini-2.5-Flash-Thinking performed lower in this task, with $mF1$ scores of $0.12$, $0.18$, and $0.17$ for $K=1, 3,$ and $6$.

Taken together, these results suggest a task-dependent effect for "thinking" tokens. They appear beneficial for complex, end-to-end localization from full images but are less advantageous when the task is simplified to few-shot classification of pre-localized objects. Overall, few-shot support improves detection over the zero-shot baseline.

\section{Discussion}
\label{sec:disc}

Our study highlights both the potential and the limitations of adapting vision-language models to the microscopic imaging domain. The results from our comprehensive evaluation on the Micro-OD benchmark yield several vital observations.

The performance gap between zero-shot and few-shot settings illustrates the critical role of in-context adaptation. While zero-shot VLMs consistently struggled due to the domain shift from natural images to microscopy, the benefit of one-shot support was limited. Most models showed little or no gain at a few shots, with the exception of the Gemini-2.5-Flash-Thinking variant, which demonstrated clear improvements. This suggests that a minimal number of in-context examples may be sufficient to anchor prior knowledge in the target domain, but the effect depends strongly on the model architecture. 

In particular, our results indicate that implicit reasoning tokens act as a form of test-time compute that is valuable when the model must integrate global spatial and semantic context. In the end-to-end few-shot detection setting (Few Shot-MMD), Gemini-2.5-Flash-Thinking substantially outperformed its non-thinking counterpart and other baselines as $K$ increased from 1 to 3 (mF1 $0.17$ to $0.21$, saturating thereafter). In contrast, non-thinking GPT-4o and GPT-5 remained near $0.05$–$0.08$ mF1. By contrast, in the hybrid setting (Few Shot-MMC), where localization is solved upstream by SAM and the VLM only classifies pre-localized crops, non-thinking models were consistently stronger (e.g., GPT-4o reached mF1 $0.30$ at $K{=}6$, versus $0.17$ for the thinking variant (\tabref{tab:results_k136}). Taken together, these findings indicate that reasoning traces are most valuable when the model must search over candidate regions and align them with language prompts; once the instance is isolated, additional reasoning may introduce redundancies, ultimately hurting performance.

\section{Conclusions}
\label{sec:con}
We present the Micro-OD benchmark and a systematic evaluation of state-of-the-art vision–language models for few-shot object detection in microscopic imagery. Our experiments demonstrate that VLMs can be adapted for improved performance in this specialized biomedical domain with only a few annotated examples, improving over zero-shot baselines. However, their effectiveness is strongly conditioned on the task design and underlying model architecture. Our analysis revealed a task-dependent effect of implicit reasoning mechanisms, where ``thinking'' tokens are more effective for complex end-to-end localization, while simpler, non-thinking models excel at fine-grained classification of pre-localized objects. Future extensions of the benchmark can include a broader range of cell types and incorporate diverse imaging modalities and acquisition settings to evaluate generalization. The Micro-OD benchmark provides a standardized foundation for these future studies, paving the way for the development of more capable AI assistants for biomedical imaging.

{
\footnotesize
\bibliography{main}
}

\appendices

\end{document}